\documentclass[11pt,a4paper]{article}
\usepackage[hyperref]{acl2020}
\usepackage{times}
\usepackage{latexsym}
\usepackage{graphicx}
\usepackage{booktabs}
\usepackage{makecell}
\usepackage{amsmath}
\usepackage{dsfont}

\usepackage{microtype}
\aclfinalcopy 

\newcommand{\minisection}[1]{\noindent{\bf #1}\hspace{0.6em}}

\title{Toward Better Storylines with Sentence-Level Language Models}

\author{Daphne Ippolito\thanks{\hspace{.5em}Work performed while a Google Student Researcher.}\\
  \texttt{daphnei@seas.upenn.edu} \\\And
  David Grangier \\
  \texttt{grangier@google.com} \\\AND
  Douglas Eck \\
  \texttt{deck@google.com} \\\And
  Chris Callison-Burch\\
  \texttt{ccb@seas.upenn.edu}}

\date{}

\begin{document}
\maketitle
\begin{abstract}
We propose a sentence-level language model which selects the next sentence in a story from a finite set of fluent alternatives. 
Since it does not need to model fluency, the sentence-level language model can focus on longer range dependencies, which are crucial for multi-sentence coherence.
Rather than dealing with individual words, our method treats the story so far as a list of pre-trained sentence embeddings and predicts an embedding for the next sentence, which is more efficient than predicting word embeddings.
Notably this allows us to consider a large number of candidates for the next sentence during training.
We demonstrate the effectiveness of our approach with state-of-the-art accuracy on the unsupervised Story Cloze task and with promising results on larger-scale next sentence prediction tasks.
\end{abstract}

\section{Introduction}

Computer generation of stories and other kinds of creative writing is a challenging endeavor. It entangles two difficult tasks: the generation of fluent natural language and the generation of a coherent storyline. In the recent year, neural language models have made tremendous progress with respect to fluency~\cite{bahdanau2014neural,vaswani2017attention,bengio2003lm,devlin2019bert}, but coherency is still a major challenge~\cite{see:2019:storyteller}. The generation of coherent stories has recently been addressed with additional conditioning: \citet{fan2018hierarchical} suggest conditioning on a story prompt, \citet{clark2018creative} propose collaboration between a generative model and a human writer, and \citet{guan2019commonsense} suggest attending to a commonsense graph relevant to the story plot. Conditioning based on a generated story plan \citep{martin2018event,fan2019strategies,yao2019plan}, a sequence of images  \citep{chandu2019storyboarding} or character roles \citep{liu2020character} have also been considered.

Our work is orthogonal to these efforts. Rather than considering additional conditioning, we propose a model which takes as input several sentences of context and selects the best next sentence within a large set of fluent candidate sentences.
We leverage pre-trained BERT embeddings \citep{devlin2019bert} to build this sentence-level language model.
Given the embeddings of the previous sentences of the story, our model learns to predict a likely embedding of the next sentence.

This task isolates the modeling of long-range dependencies from the prediction of individual words, which has several advantages.
First, since our model only needs to determine how well each candidate sentence would fit as a coherent continuation to the story, it does not spend capacity and time to learn fluency.
Second, our model does not manipulate individual words but full sentences, which allows us to consider tens of thousands of candidate sentences at a time.
This contrasts with prior work \citep{logeswaran2018an} where the need to learn token-level representations limited the number of candidate next sentences that could be considered to a few hundred.
Third, we can rely on compact model architectures that train quickly because we take advantage of strong semantic representation from a pre-trained bidirectional language model, BERT, as our sentence embeddings.
Of course, these benefits also imply that our sentence representation is limited to the information extracted by the pre-trained model.
Nevertheless, we show that our model achieves state-of-the-art accuracy among unsupervised approaches on the Story Cloze task: predicting which of two sentences coherently ends a short story.

Our work also opens up the possibility of ranking thousands of candidate sentences from a large literature repository.
On the ROC Stories dataset, we observe that training with a large number of candidates is key for selecting the most coherent ending among a large set of candidates at test time.
We also show preliminary results on the efficacy of our method for ranking candidate next sentence on the Toronto Book Corpus~\citep{kiros2015skip}, a much larger book dataset.
We envision that our methods for scoring many candidate next sentences by their coherence with the context might be useful to downstream generation tasks where it is possible to generate many fluent continuations of a text, but it remains an unsolved problem how to refine and choose the best of them. To encourage this exploration, we release our code and models\footnote{Code for ROC Stories experiments can be found at {\small \url{https://github.com/google-research/google-research/tree/master/better_storylines}}.}.

\section{Proposed Method}

We propose a sentence-level language model: our model estimates $P(s_{t+1} | s_{1:t})$, the probability distribution for sentence $s_{t+1}$ given the $t$ previous sentences, $s_1, \ldots s_t$. Since it is intractable to marginalize over all possible candidate next sentences, we consider a finite but large set of $N$ valid, fluent sentences. Without loss of generality, we can consider $s_{t+1} \in \{1, \ldots, N\}$ as an integer index into that set of possible next sentences. This strategy resembles negative sampling in word2vec~\cite{mikolov:word2vec}.

Our model represents sentences with pre-computed vector embeddings. Specifically, sentences are represented by the mean of the 768-dimensional contextual word embeddings of the second-to-last layer of BERT~\citep{devlin2019bert}. This representation has shown to encode more transferable features compare to other layers~\citep{liu2019linguistic}. Alternative sentence representations were considered, including embeddings from the universal sentence encoder \citep{cer2018universal} and a weighted mean of the BERT embeddings using inverse document frequency weighting \citep{zhang2019bertscore}. None of these alternatives improved our results however.

Motivated by simplicity, we consider a classical multi-layer perceptron (MLP) $f_\theta$ which takes as input the context sentence embeddings concatenated into a single vector. At the output layer, we perform a softmax operation. If we represent candidate sentences $\{1, \ldots, N\}$ by the embeddings $\{e_i\}^N_{i=1}$, our model estimates the probability that $i$ is the next sentence by the softmax
\begin{align*}
\log P(s_{t+1} = i|s_{1:t}) = e_i^\top h - \log Z(h) 
\end{align*}
where $h = f_\theta(s_{1:t})$ is the output of the MLP given context $s_{1:t}$, and $Z(h) = \sum_{j=1}^{N} \exp e_j^\top h$ is the partition function. At train time, the candidate set $\{1, \ldots, N\}$ consists of the correct next sentence along with $N-1$ distractor sentences. The distractors can either be static (the same set used throughout training) or dynamic (picked at random from a larger set for each train batch). In this case, the ``vocabulary" of next values to choose from changes with each train step, similar to negative sampling~\cite{mikolov:word2vec}.
At test time, novel sentences can be embedded with BERT and scored by our model.

Like a classical language model, we optimize for the likelihood of the true next sentence's embedding.
However, when training we found that the sentences from the context ($s_1,\ldots,s_t$) often ended up being given very high scores by our model. Inspired by work in sentence reordering \citep{lapata2003ordering,logeswaran2018an}, we incorporated an auxiliary loss, which we refer to as \textbf{CSLoss}, that only includes the context sentences $s_{1:t}$ in the distractor set.

Lastly, we consider a residual variant of the MLP (referred to as \textbf{resMLP}) with skip connection between layers, as described in \citet{he2016deep}.
The residual model trains faster and sometimes achieves higher accuracy than the non-residual model.
Though we experimented with recurrent \citep{sundermeyer2012lstm} and self-attention \citep{vaswani2017attention} models, we did not observe improvements, perhaps because the input to our model is already the high-dimensional output of a large mask language model.
We leave deeper architecture exploration, which will be especially critical as context length is extended, to future work.

\section{Experimental Setup}

We first describe our experiments on the ROC Stories dataset of short 5-sentence stories before showing our setup on the larger Toronto Book Corpus.


\begin{table*}[h]
  \centering
  \small
    \begin{tabular}{l l||r|r|r|r}
              &           & Valid 2016 & Test 2016 & Valid 2018 & Test 2018\\
    \hline
    Our model & MLP       & 69.7 & 68.8 & 70.1 & 69.0 \\
              & + CSLoss  & \textbf{73.5} & \textbf{73.0} & \textbf{73.1} & \textbf{72.1} \\
    \hline
    Alternatives & \citet{peng2017joint} & --  & 62.3 &  -- &  --\\
                 & \citet{schenk2017resource}  & 62.9 & 63.2 &  -- &  --\\
    \hline
    Lang. Models & \citet{schwartz2017story} & -- & 67.7 & -- &  --\\
        & GPT-2 \citep{radford2019lm} & 54.5 & 55.4 & 53.8 &  --\\
        & GPT-2 + finetuning          & 59.0 & 59.9 & 59.0 &  --\\
    \end{tabular}%
    \caption{Accuracies (\%) for the Story Cloze binary classification task. \citet{schwartz2017story} is a semi-supervised technique. GPT-2 refers to predicting the more likely ending according to the 355M parameter model, and GPT-2 finetuning was done on the ROC Stories train set.}
  \label{tab:roc_task_eval}%
\end{table*}%

\subsection{ROC Stories}

\minisection{Dataset} Our experiments use the ROC Stories dataset, which consists of stories focusing on common sense \cite{mostafazadeh2016rocstories}.
The training set has 98k stories, with five sentences each. The validation and test sets each contain 1.8k stories consisting of four sentences followed by two alternative endings: one ending is coherent with the context; the other is not.
The dataset was introduced for the Story Cloze task, inspired by ~\citet{taylor1953cloze}, where the goal is to select the coherent ending.
While the dataset and task were introduced as a way to probe for coherence and commonsense in models trained only on the unlabeled portion, most research
derived from this dataset focuses on a supervised setting, using the validation set as a smaller, labeled training set \citep{chaturvedi2017story,sun2019reading,cui2019story,li2019story,zhou2019story}.
Our work is faithful to the original task objective. We train solely on the training set, i.e. the model never sees incoherent endings at training time.

\minisection{Model} We consider two models, an MLP and a residual MLP. They take as input the previous sentences represented as the concatenation of their embeddings. 
Alternative context aggregation strategies were considered with recurrent \citep{sundermeyer2012lstm} and attention \citep{vaswani2017attention} architectures, without strong empirical advantages.
The models maps its input to a vector which is compared to a set of candidate sentence embeddings via dot product. The embedding of the true next sentence should receive the highest score. For each example, we consider all other fifth sentences in the training set (96k in total) as the candidate set. 

The input of our model is 3,072 dimensional, i.e. 4 context sentences represented by 768 dimensional BERT embeddings. After an architecture search, our best MLP has 3 layers of 1,024 units, and our best resMLP has a single residual layer with hidden size of 1,024. Both contain just over 6M trainable parameters.
Both apply dropout with a rate of 0.5 after each ReLU, and layer normalization is performed on the concatenated context sentence embedding passed in as input to the network and on the final predicted embedding for the next sentence.
For the Story Cloze task, the two architectures achieve similar validation accuracy, but when considering more than two distractors, the resMLP significantly outperforms the standard MLP.
The resMLP also converges quicker than the MLP.
Training to convergence takes under 2 hours for each model on a Tesla V100.

\subsection{Toronto Book Corpus}

\minisection{Dataset} ROC Stories contains only self-contained five-sentence stories, focusing on everyday life scenarios. They contain no dialog and very little flowery, expository language. Ideally our method would also be successful at scoring potential continuations to more naturally-written stories. To this end, we test out our approach on excerpts from the Toronto Book Corpus \citep{kiros2015skip}, a dataset of self-published novels.
The dataset contains over 7,000 unique books totalling over 45 million sentences. Since these stories are much longer than the ROC Stories ones and many of the sentences are uninformative (nearly 5\% of sentences are 3 words or shorter, and 14\% are 5 words or shorter), we double the context length to 8 sentences.

\minisection{Model} In addition to experimenting with a similar residual MLP architecture to the one used on ROC Stories, we also ran experiments with a Transformer model \citep{vaswani2017attention}. The residual MLP architecture contains 2 residual layers with hidden size of 1024 (11M params total). 
The transformer has 4 self-attention layers with hidden size of 768, filter size of 2048 and 8 attention heads (22M params total).
While the residual MLP is trained to predict the 9th sentence given the previous 8 sentences, the Transformer is trained to predict each next sentence given the previous sentences in a sequence of length 10 sentences.
However, we only evaluate the Transformer on the task of predicting the 9th sentence so that evaluation results are directly comparable to the residual MLP.

For each batch during training, 2k distractors are randomly selected from the train set.
Like with ROC Stories, we experiment with an auxiliary loss where just sentences from the context were used as distractors.
Table~\ref{tab:full_task_eval} reports the results.


\begin{table}[]
  \centering
  \small
    \begin{tabular}{l||rr}
     & P@10 & MRR \\
    \hline
    MLP   & 6.2 & 0.052  \\
    +CSLoss   & 3.4 & 0.029 \\
    \hline
    ResMLP & \textbf{10.3} & \textbf{0.087} \\
    +CSLoss &  6.2 & 0.051 \\
    \hline
    Random & 0.01 & 2e-5 \\
    \end{tabular}%
  \caption{Precision@10 and mean-reciprocal rank on the 2018 valid set when considering all 5th sentences in the train and valid sets (98k total) as candidate endings.}
  \label{tab:full_task_eval}%
\end{table}%

\section{Results}
We evaluate on the Story Cloze task, a binary classification task, as well as on the task of ranking a large set of possible next sentences.

\subsection{Story Cloze Task} 

Table \ref{tab:roc_task_eval} shows that our method outperforms unsupervised alternatives. The introduction of the CSLoss which considers only context sentences as candidates improves accuracy compared to only using a loss over all possible fifth sentences.

For comparison, we include the accuracies of the best unsupervised methods in the literature. \citet{schenk2017resource} construct negative examples for their binary classification task by pairing contexts with random fifth sentences selected from the training set. \citet{peng2017joint} train a language model to predict a representation of the semantic frame, entities, and sentiment of the fifth sentence given the representations of the previous sentences, then take the more likely fifth sentence. We achieve higher accuracy without relying on a task-specific architecture.

Table \ref{tab:roc_task_eval} also shows that picking the ending that is more likely according to a word-level language model, in our case GPT-2's 355M parameter model, does not yield very high accuracies, even when the language model is finetuned on ROC Stories text~\citep{radford2019lm}.
Lastly, we also include the accuracy reported by \citet{schwartz2017story}, where a logistic classifier is trained to combine multiple language model scores.

It is worth noting that state-of-the-art on the Story Cloze task is over 90\% accuracy \citep{li2019story,cui2019story} for semi-supervised settings. The methods achieving this level of performance are not comparable to our {\it unsupervised} approach as they require training on the {\it labeled} validation set. The language model approach from \citet{schwartz2017story} also falls into this category.

\begin{figure}
\centering
\includegraphics[width=0.5\textwidth]{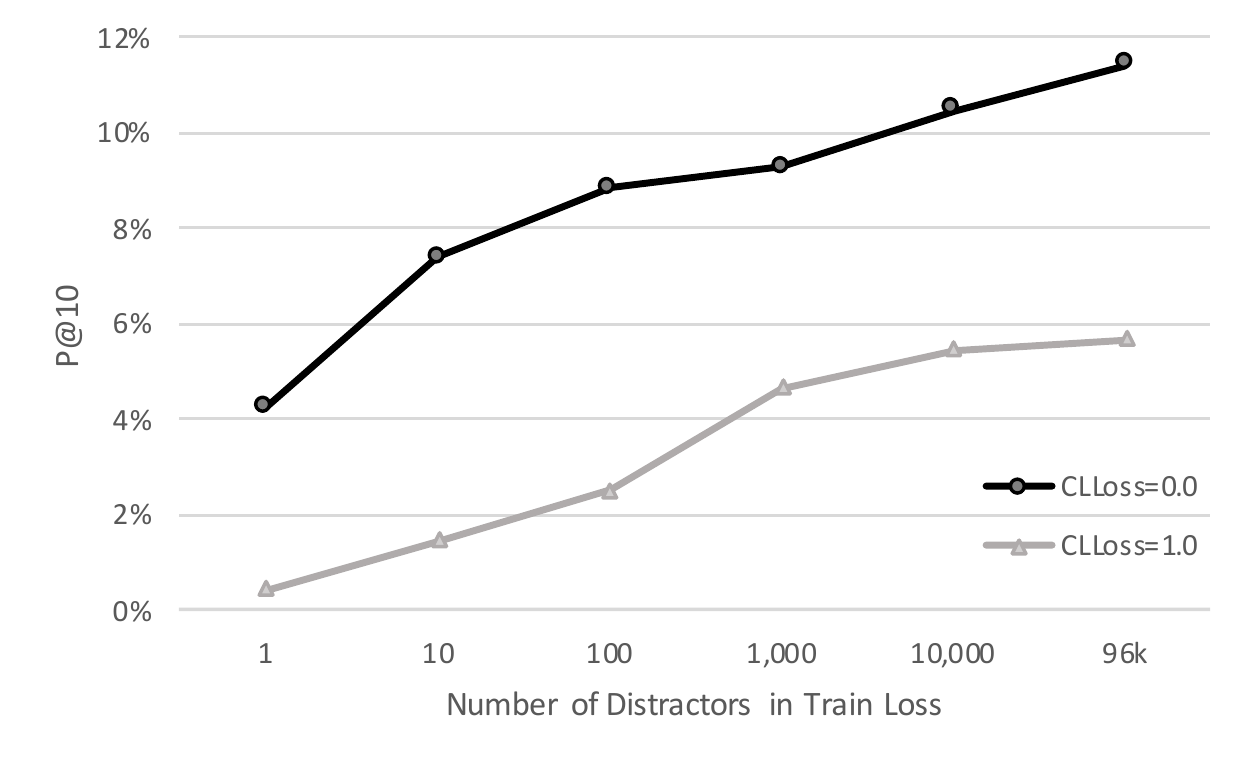}
\caption{The impact of the number of negative sentences used during training on the rank of the true ending out of 98k distractors. Results are with the resMLP on the 2018 valid set.}
\label{fig:num_distractors}
\end{figure}

\subsection{Ranking Many Sentences on ROC Stories} 
For generation and suggestion scenarios, it is useful to be able to surface the best next sentence out of hundreds or thousands of candidates. In Table \ref{tab:full_task_eval}, we show the performance of our method on the 2018 validation set when all 98,161 fifth sentences in the training set plus all 1,571 correct 5th sentences in the 2018 validation are considered as candidate endings.
Top-10 accuracy is highest, at 10.3\%, when training a residual MLP without CSLoss.

Interestingly, strong performance on the Story Cloze task does not necessarily translate to strong performance on the large-scale ranking task.
The CSLoss improves performance on the Story Cloze task but hurts it for large-scale ranking.

In Figure \ref{fig:num_distractors}, we show how large-scale ranking performance improves as the size of the train-time distractor set is increased.
However, on the Story Cloze task, the number of training distractors has no significant impact on performance.
Even when only a single distractor is randomly chosen at each step of training, our method achieves over 70\% 2016 test accuracy.
It seems that training for the goal of detecting the true next sentence out of a very diverse candidate set is useful at test time only when the set of distractors at test time is similarly large and diverse.
The many-distractors training regime might be less useful for the Story Cloze task since the two candidate endings are designed to be quite topically similar to each other.

Some qualitative examples are shown in Table \ref{tab:qualEval}.
The failure examples showcase a side-effect of relying on pre-trained sentence embeddings: if common names like ``Becky" or ``Laura" or sports such as ``fishing" and ``golf" are close to each other in embedding space, our model will fail to distinguish between them.


\noindent
\begin{table}[]
  \centering
  \small
    \begin{tabular}{l|rrr}
        & 10k & 100k & same book \\
        \hline
        resMLP & 22.5\% & 7.4\% & 7.8\% \\
        +CSLoss & 11.5\% & 2.5\% & 5.3\% \\
        \hline
        Transformer & 15.2\% & 	4.0\% & 4.8\% \\
        +CSLoss & 4.8\% & 0.8\% & 2.0\% \\
    \end{tabular}%
  \caption{Precision@10 On Toronto Book Corpus for retrieving the correct next sentence (given the 8 previous sentences) when considering 10k or 100k distractor sentences, or all of the sentences from the same book as distractors.}
  \label{tab:full_task_eval}%
\end{table}%

\begin{table}[t]
  \centering
  \tiny
    \begin{tabular}{p{31em}}
    \Xhline{1.5pt}
    \textbf{Context:} My family got up one morning while on vacation. We loaded our boat onto a trailer and drove to the beach. After loading up from the dock, we took off on our boat. After only a few minutes on the sea, dolphins began to swim by us. \\
    \hline
    \textbf{GT:} (22.89) We played with them for a while and then returned to the dock. \\
    \textbf{Rank:} 9\\
    \hline
    \multicolumn{1}{l}{\textbf{Top scored:}} \\
    (25.06)  We were definitely lucky to see them and it made the trip more fun! \\
    (24.31)  They loved everything about that trip and vowed to do it again! \\
    (23.76)  We were sad to come home but excited to plan our next vacation. \\
    (23.72)  It was one of our best vacations ever! \\
    \Xhline{1.5pt}
    \textbf{Context:} Ellen wanted to be smart. She started reading the dictionary. She learned two hundred new words the first day. Ellen felt smart and educated. \\
    \hline
    \textbf{GT:} (30.23) She couldn't wait to use the new words. \\
    \textbf{Rank:} 1\\
    \hline
    \multicolumn{1}{l}{\textbf{Top scored:}} \\
    (30.23)  She couldn't wait to use the new words. \\
    (29.78)  She felt like a new woman when she was done! \\
    (29.01)  She decided to go back to speaking like her normal self! \\
    (28.95)  She felt like a new girl! \\
    \Xhline{1.5pt}
    \textbf{Context:} It was a very cold night. Becky was shivering from the cold air. She needed to cover up before she caught a cold. She wrapped up in her favorite blanket. \\
    \hline
    \textbf{GT:} (18.717398) Becky finally got warm. \\
    \textbf{Rank:} 3,028\\
    \hline
    \multicolumn{1}{l}{\textbf{Top scores:}} \\
    (39.09)  Laura ended up shivering, wrapped in a blanket for hours. \\
    (36.71)  After being cold all day, the warmth felt so good. \\
    (33.77)  Sam was able to bundle up and stay cozy all winter. \\
    (33.38)  The breeze felt good on her wet shirt. \\
    \Xhline{1.5pt}
    \textbf{Context:} Benjamin enjoyed going fishing with his grandfather as a kid. They would pick a new location to go to every summer. Benjamin liked seeing who would catch the biggest fish. Even after his grandfather passed he continued the tradition. \\
    \hline
    \textbf{GT:} (26.65) He now takes his own grandchildren to create memories for themselves. \\
    \textbf{Rank:} 2,281\\
    \hline
    \multicolumn{1}{l}{\textbf{Top ranked:}} \\
    (34.71)  Greg grew to love golfing and is now his favorite thing to do. \\
    (33.82)  It was a tradition Tim continues with his own family. \\
    (33.63)  Alex learned to be grateful of his family's unique tradition. \\
    (33.40)  Tom was sad that he would have to let his son down. \\
    \Xhline{1.5pt}
    \end{tabular}%
  \caption{Top-scoring sentences (using resMLP without CSLoss) among 98k possible endings when using prompts from the validation set. Two success and two failures cases are shown.}
  \label{tab:qualEval}%
\end{table}%

\subsection{Ranking Many Sentences on Toronto Book Corpus}

When evaluating with 100k distractors, about as many as our ROC Stories large-scale ranking task, P@10 is at best 7.1\%, compared with 22.7\% for ROC Stories.
We suspect that this task would benefit from longer contexts and better selection of distractors.
In particular, a qualitative evaluation of the data highlighted the presence of a large quantify of short, generic sentences in the high ranking sentences (e.g. ``he said." and ``Yes.").
We see reducing the density of such sentences at training time as a potential for improvement.

In addition, further investigation is necessary into why the Transformer did not work as well as the residual MLP.
The use of variable sequence length architectures like the Transformer will become more critical as the input sequence length is increased beyond what an MLP can easily handle.

\section{Conclusions}

This work introduces a sentence-level language model which takes a sequence of sentences as context and predicts a distribution over a finite set of candidate next sentences. It takes advantage of pre-trained BERT embeddings to avoid having to learn token-level fluency, allowing the model to focus solely on the coherence of the sentence sequences. Our results on the Story Cloze task highlight the advantage of this strategy over word-level language models. At train time, our model considers much larger amounts of text per update than typical token-level language models. We show that this strategy allows our model to surface appropriate endings to short stories out of a large set of candidates.

As future work, we plan to further evaluate the impact of different sequential architectures, longer contexts, alternative sentence embeddings, and cleverer selection of distractors.
Inspired by deliberation networks and automatic post editing methods \citep{xia2017deliberation,freitag2019ape}, we ultimately want to apply our model to two-step generation, first selecting a sentence from a large set before refining it to fit the context.

\section*{Acknowledgements}

This research is based upon work supported in part by U.S. DARPA KAIROS Program No. FA8750-19-2-1004. The views and conclusions contained herein are those of the authors and should not be interpreted as necessarily representing the official policies, either expressed or implied, of DARPA or the U.S. Government. The U.S. Government is authorized to reproduce and distribute reprints for governmental purposes notwithstanding any copyright annotation therein.

\bibliography{acl2020}
\bibliographystyle{acl_natbib}

\end{document}